\documentclass[sigconf]{acmart}
\AtBeginDocument{%
  \providecommand\BibTeX{{%
    \normalfont B\kern-0.5em{\scshape i\kern-0.25em b}\kern-0.8em\TeX}}}

\usepackage[linesnumbered,ruled,vlined,algo2e]{algorithm2e}
\usepackage{multirow}
\usepackage{tabularx}
\usepackage{graphicx}
\usepackage{float}
\usepackage{subfigure}
\usepackage{array}
\usepackage{amsfonts}
\usepackage{epsfig}
\usepackage{graphicx}
\usepackage{amsmath}
\usepackage{booktabs}
\usepackage{appendix}
\usepackage{bbm}
\usepackage{stfloats}

\usepackage{amsthm}
\usepackage{mathtools}
\usepackage{commath}
\usepackage[normalem]{ulem}

\newcommand{\PreserveBackslash}[1]{\let\temp=\\#1\let\\=\temp}
\newcolumntype{C}[1]{>{\PreserveBackslash\centering}p{#1}}
\newcolumntype{R}[1]{>{\PreserveBackslash\raggedleft}p{#1}}
\newcolumntype{L}[1]{>{\PreserveBackslash\raggedright}p{#1}}

\usepackage{enumitem}
\setlist[itemize]{leftmargin=*}

\newcommand{\ours}{ACT\xspace}

\newcommand{\diffranker}{$\text{R}^2$\xspace}
\newcommand{\ourmodel}{$\text{ACTR}^2$}
\newcommand{\ourmodels}{$\text{ACTR}^2$\xspace}

\newtheorem{problem}{Problem}
\newtheorem{proposition}{Proposition}

\usepackage{todonotes} 

\usepackage{amsmath}
\begin{abstract}
Recent automated machine learning systems often use learning curves ranking models to inform decisions about when to stop unpromising trials and identify better model configurations.
In this paper, we present a novel learning curve ranking model specifically tailored for ranking normalized entropy (NE) learning curves, which are commonly used in online advertising and recommendation systems. Our proposed model, self-\underline{A}daptive \underline{C}urve \underline{T}ransformation augmented \underline{R}elative curve \underline{R}anking (\ourmodels), features an adaptive curve transformation layer that transforms raw lifetime NE curves into composite window NE curves with the window sizes adaptively optimized based on both the position on the learning curve and the curve's dynamics. We also introduce a novel differentiable indexing method for the proposed adaptive curve transformation, which allows gradients with respect to the discrete indices to flow freely through the curve transformation layer, enabling the learned window sizes to be updated flexibly during training. Additionally, we propose a pairwise curve ranking architecture that directly models the difference between the two learning curves and is better at capturing subtle changes in relative performance that may not be evident when modeling each curve individually as the existing approaches did. Our extensive experiments on a real-world NE curve dataset demonstrate the effectiveness of each key component of \ourmodels and its improved performance over the state-of-the-art.

\end{abstract}


\begin{document}
\pagestyle{plain}
\title{Learning to Rank Normalized Entropy Curves with Differentiable Window Transformation}


\author{Hanyang Liu}
\affiliation{%
  \institution{Washington University in St. Louis}
  \streetaddress{1 Brooking Drive}
  \city{St. Louis}
  \country{United States}
  \postcode{63130}
}
\email{hanyang.liu@wustl.edu}

\author{Shuai Yang}
\affiliation{%
  \institution{Meta}
  \streetaddress{One Hacker Way}
  \city{Menlo Park}
  \country{United States}
  \postcode{94025}
}
\email{shuaiyang@meta.com}

\author{Feng Qi}
\affiliation{%
  \institution{Meta}
  \streetaddress{One Hacker Way}
  \city{Menlo Park}
  \country{United States}
  \postcode{94025}
}
\email{fqi@meta.com}

\author{Shuaiwen Wang}
\affiliation{%
  \institution{Meta}
  \streetaddress{One Hacker Way}
  \city{Menlo Park}
  \country{United States}
  \postcode{94025}
}
\email{sw2853@meta.com}


\renewcommand{\shortauthors}{Hanyang Liu et al.}

\maketitle


\section{Introduction}
Learning curves visualize how a machine learning (ML) system's performance (e.g., loss, accuracy) evolves over time or with an increasing number of training examples during an iterative optimization process. \textit{Normalized entropy} (NE)~\cite{he2014practical} curves, as a common type of learning curves, are widely used to continuously evaluate the performance of ML models in the field of online advertising and recommendation systems~\cite{qu2016product,hazelwood2018applied,aryafar2017ensemble,du2021alternate,qin2020user}. 
Learning curves, such as NE curves, have been serving as a crucial indicator in determining when to stop training an ML model before the iterative training process reaches a previously set budget, i.e., early termination~\cite{wistuba2020learning,domhan2015speeding,chandrashekaran2017speeding,wang2021rank,jiang2022neural}. In addition to the most common practice of using it to prevent overfitting (i.e., early stopping~\cite{caruana2000overfitting,prechelt1998early}), early termination has also been used to speed up the automated optimization (i.e., autotuning) of neural architectures or hyperparameters~\cite{elsken2019neural,wistuba2020learning,mohr2021towards,mohr2021fast,swersky2014freeze} by stopping the unpromising trials early.


To inform decisions about when to stop training and identify more promising configurations, prior works ~\cite{chandrashekaran2017speeding,domhan2015speeding,klein2016learning,baker2017accelerating} rely on curve extrapolation to predict the final performance of a model configuration based on its partially observed learning curve. This way of determining the superiority of different curves is known as \textit{pointwise ranking}~\cite{liu2009learning} and has been proven to be less efficient than \textit{pairwise ranking} methods~\cite{burges2005learning}, which directly optimize for the relative superiority.
Therefore, recent works~\cite{wistuba2020learning,wang2021rank} have taken a different approach by formulating the decision-making process for early termination as a pairwise ranking problem which focuses
on relative performance.
LCRankNet~\cite{wistuba2020learning}, which has a Siamese structure~\cite{burges2005learning,doughty2018s,wang2021rank} with each twin network component modeling a curve separately, is used to predict the final superiority of a pair of learning curves. It has consistently outperformed pointwise methods and is considered the state-of-the-art in learning curve ranking. It has been shown that, focusing on relative performance, rather than absolute performance, can help generalize knowledge across domains and increase the sample efficiency of learning curves, as the range of individual model performance can vary significantly depending on the domain (e.g., model type, task, dataset). However, we argue that all the above methods face two major limitations that may lead to suboptimal results.

Firstly, the approaches mentioned above do not take into account the unique characteristics of different types of learning curves, treating all curves equally. This can result in suboptimal ranking performance, since the specific features of a particular type of learning curve can be useful in improving ranking results. In the context of online advertising and recommendation systems, there are two types of NE curves, \textit{lifetime NE} (LNE) and \textit{window NE} (WNE). LNE is used to evaluate the overall model performance throughout the training process, while WNE is used to evaluate the model's latest performance on the most recent examples within a specific time window. Typically, LNE is observed and recorded as the learning curve, but the end-point WNE value with a specific window (e.g., 1-day WNE) is the de facto criterion to determine which model configuration is superior. Ignoring the different characteristics of NE curves and using raw LNE values as input can lead to a discrepancy between the input and targeted outcome, resulting in unreliable results. 
Simply replacing the LNE curve with a transformed WNE curve of a specific window size is still suboptimal, as both LNE and WNE curves may contain predictive information with different temporal ranges that indicate the final superiority of model configurations. Additionally, the dynamics of different curves and different positions on the same curve vary, making it difficult to find a single window size that fits for all cases.

Secondly, Siamese-based ranking approaches, including the state-of-the-art ranking model LCRankNet~\cite{wistuba2020learning}, may struggle to accurately rank curves with small differences, especially if the differences are inconsistent over time. 
This is because, while these methods optimize pairwise ranking for relative superiority, they still implicitly model each curve individually using the twin network components in the Siamese structure.
However, these components may not be sensitive enough to capture subtle relative variations between curves because they focus on modeling the full dynamics and temporality of each curve individually, and not all of these patterns may be relevant to the relative superiority of the curves. For example, two curves going up and down concurrently may reflect variations in their expected final performance, but do not necessarily indicate changes
in their relative superiority. This limitation can be especially problematic for tuning large and advanced models in the industry, as their training often converges with small differences in performance. For example, even a 0.05\% difference in NE is considered significant for large-scale online advertising models.
In contrast, we propose that the dynamics and temporality of the difference between two curves are better indicators of final superiority than those of each curve individually, since the relative dynamics is what ultimately determines which curve is superior.

To address both the two limitations, we propose a novel pairwise learning curve ranking model, self-\underline{A}daptive \underline{C}urve \underline{T}ransformation augmented \underline{R}elative curve \underline{R}anking (\ourmodels), specifically for the application in ranking NE curves widely used in the computing industry. 
To the best of our knowledge, \ourmodels is the first curve ranking model specifically tailored for models with NE learning curves.
\ourmodels features the following key components:
\begin{itemize}
    \item A self-adaptive curve transformation (ACT) layer that transforms raw LNE curves into composite WNE curves with window sizes optimally adapting to the position on the learning curve and the curve's dynamics. The composite WNE curve is optimized for improved ranking results and can be considered the integration of a LNE curve and a spectrum of WNE curves with various windows. Importantly, the ACT layer is orthogonal to existing approaches and can be used in conjunction with any curve ranking model to improve performance.
    \item A novel differentiable indexing mechanism specifically for the proposed ACT layer based on a soft one-hot reparameterization trick to allow gradients with respect to the discrete indices to flow freely through the ACT layer, enabling the learned window sizes to be updated flexibly during training.
    \item A novel pairwise curve ranking architecture that is based on the relative values of the curves and directly models the difference between the two learning curves. This design is intended to better capture changes in relative performance that may not be evident in the individual curves themselves.
    \item Extensive experiments on a real-world NE curve dataset collected from a major autotuning platform in the industry demonstrate the effectiveness of each key component of \ourmodels and the significant improvement over the state-of-the-art.
\end{itemize}

\section{Preliminaries}

\subsection{Problem Statement}\label{sec:problem}
Formally, we formulate learning curve ranking as the following pairwise learning-to-rank problem:
\begin{problem}\label{problem}
Given a pair of models with different model configurations, $\{m_i, m_j\} \in \mathcal{M}$, with their corresponding descriptive attributes $\{\mathbf{x}_i, \mathbf{x}_j\}$ (e.g., architecture, hyperparameter) and  partially observed NE learning curves $\{\mathbf{y}_i, \mathbf{y}_j\}$ where $\mathbf{y} = [y_1, y_2, ..., y_L]^T$ has $L$ measures at regular intervals, our goal is to train a pairwise ranking model $g(\cdot)$ that predicts the final superiority of $m_i$ and $m_j$ when the training is finished by estimating the probability that $m_i$ is better than $m_j$, $p(m_i > m_j) = g((\mathbf{x}_i, \mathbf{y}_i), (\mathbf{x}_j, \mathbf{y}_j))$.
\end{problem}


\subsection{Related Work}
Most of the prior work related to learning curve ranking has focused on the pointwise prediction of final individual model performance using partial learning curves. Prior efforts~\cite{jamieson2016non,li2017hyperband} simply used the last observed point on the learning curve to determine when to stop unpromising trials. Mohr et al~\cite{mohr2021towards,mohr2021fast} extrapolated the curve based on the slope of the last two observed points and used a fixed threshold to make termination decisions. Others~\cite{domhan2015speeding,chandrashekaran2017speeding,klein2016learning} used an ensemble of parametric formulas to estimate performance based on partial learning curves, with the parameters calibrated to fit the observed curves. 
Without relying on those basic formulas, Baker et al~\cite{baker2017accelerating} proposed to directly forecast the final performance based on extracted features of the observed curves using a regressive model. All these pointwise approaches rely on curve extrapolation to determine relative superiority and tend to introduce more errors during the extra step. In addition, pointwise ranking has been proven to be less efficient than pairwise ranking that directly optimize for relative superiority~\cite{burges2005learning}. 
Recently, Wistuba et al~\cite{wistuba2020learning} proposed to formulate learning curve ranking as a pairwise ranking problem, and its underline model has achieved state-of-the-art performance in ranking partial learning curves from models across domains. However, the Siamese ranking structure used in~\cite{wistuba2020learning} is not well-suited for ranking curves with small differences, especially when those differences are inconsistent throughout the training process.

In contrast to these previous approaches, our work focuses on directly modeling the difference between learning curves in order to accurately capture subtle patterns of relative change. While building upon pairwise ranking, our proposed framework is further tailored for the unique characteristics of NE curves that are widely used in the computing industry. The proposed adaptive curve transformation scheme is complementary to the previous approaches and can be used in conjunction with them.

\subsection{Pairwise Curve Ranking}\label{sec:ranking}

\subsubsection{Pairwise Ranking Loss}
A natural solution to pairwise ranking is to train a model to evaluate the relative performance, \textit{distance}, of the two candidates, $\delta_{i,j}\in\mathbb{R}$, where a larger distance reflects less similar the two candidates are.
Distance is directional, i.e., $\delta_{i,j} > 0$ if $m_i > m_j$, and $\delta_{i,j} \leq 0$ if $m_i \leq m_j$. Then the probability that candidate $m_i$ is better than candidate $m_j$ is defined in~\cite{burges2005learning} as
\begin{equation}\label{eq:prob}
    p(m_i > m_j) = \hat{p}_{i,j} = \frac{e^{\delta_{i,j}}}{1 + e^{\delta_{i,j}}}
\end{equation}
To train this model, the above estimated value $\hat{p}_{i,j}$ is mapped to match the pairwise binary labels $p_{i,j}$ defined as
\begin{equation}
    p_{i,j} = 
    \begin{cases}
      1 & \text{if\ } m_i \geq m_j\\
      0 & \text{if\ } m_i < m_j
    \end{cases}
\end{equation}
The model parameters are optimized by minimizing the cross-entropy loss over all arbitrary model pairs $m_i$ and $m_j$
\begin{equation}\label{eq:ce_loss}
    \mathcal{L}_{\text{CE}} = \sum_{i,j}-p_{i,j}\log{\hat{p}_{i,j}} - (1 - p_{i,j})\log{(1 - \hat{p}_{i,j})}
\end{equation}

\subsubsection{LCRankNet: Pairwise Curve Ranking via Siamese Network}
\label{sec:lcranknet}
To solve Problem \ref{problem}, the prior work \cite{wistuba2020learning} proposed to adopt a Siamese neural network~\cite{burges2005learning} as the ranking model, called LCRankNet. In LCRankNet, each of the two twin neural networks (sharing all parameters), $f_{\boldsymbol{\theta}}: \mathbb{R}^D\rightarrow\mathbb{R}$, serves as a scoring function to predict the final performance of each curve, where $\boldsymbol{\theta}$ denotes the trainable parameters of $f$. The scoring function takes the model characteristics $\mathbf{x}$ and the partial learning curve $\mathbf{y}$ as input, and the distance is defined as the subtraction of the two scoring function outputs
\begin{equation}\label{eq:delta}
    \delta_{i,j} = f_{\boldsymbol{\theta}}(\mathbf{x}_i, \mathbf{y}_i) - f_{\boldsymbol{\theta}}(\mathbf{x}_j, \mathbf{y}_j)
\end{equation}
A specific scoring model $f$ is trained on all partial learning curves of length $L$ to predict the probabilities of curves of length $L$.


The scoring function $f(\cdot)$ in LCRankNet~\cite{wistuba2020learning} takes into consideration both the partially observed learning curves and the characteristics of the model (i.e., metadata), including hyperparameters,  network architectures, etc., as input. It is composed of several sub-networks, each of which serves as an encoder to transform a portion of the model information into a vector representation. 
The outputs of each sub-networks are then concatenated
and directly passed to a fully connected (FC) layer with a scaler
output as the score.

\subsection{Normalized Entropy Curve}\label{sec:ne}
\subsubsection{Normalized Entropy}
Normalized entropy (NE), or formally, normalized cross-entropy, is defined as the average log loss per impression (an impression~\cite{he2014practical} is counted each time the ad is shown on a page) of the evaluated model on a specific dataset divided by that of a baseline model as the normalization factor. This baseline model produces the label randomly so that the label distribution follows the background click through rate (CTR), which is calculated using the average empirical CTR of the training dataset~\cite{he2014practical}. 
Assume a give training dataset has $N$ examples with labels $q_i\in\{0, 1\}$, the model estimated probability of click $\hat{q}_i$ where $i=1, 2, ..., N$, and the average empirical CTR on the whole $N$ examples as $\Bar{q}$, NE is defined formally as
\begin{equation}\label{eq:NE}
    \text{NE} = \frac{\frac{1}{N}\sum_{i=1}^{N}l(q_i, \hat{q}_i)}{l(\Bar{q}, \Bar{q})}
\end{equation}
where 
\begin{equation}\label{eq:loss}
    l(a, b) = -a\log(b) - (1 - a)\log(1 - b)
\end{equation}
Using the log loss of the background CTR, $l(\Bar{q}, \Bar{q})$, to normalize the average cross-entropy makes NE of the evaluated model insensitive to datasets with intrinsically different CTR. Hence model comparison across datasets is viable using NE. 

\subsubsection{Observation Curve}
In online advertising and recommendation, it is common to train a model with unrepeated continuously streaming data examples (i.e., all examples are only used once for training and there is only one training epoch in total). 
To closely monitor the training process, an \textit{observation curve} is used to evaluate the evolving average model performance over the cumulative amount of training examples (e.g., loss vs. number of examples)~\cite{mohr2022learning}. 
During the training process, a new batch of training examples is used to evaluate the current model before the model parameters are updated using the same batch of data.
In contrast, in many other scenarios~\cite{wistuba2020learning,domhan2015speeding}, an \textit{iteration curve} is often used to describe how model performance on a fixed set of examples evolves over iterations (e.g., loss/accuracy vs. number of epochs).

\begin{figure}[t]
    \centering
    \includegraphics[width=0.9\linewidth]{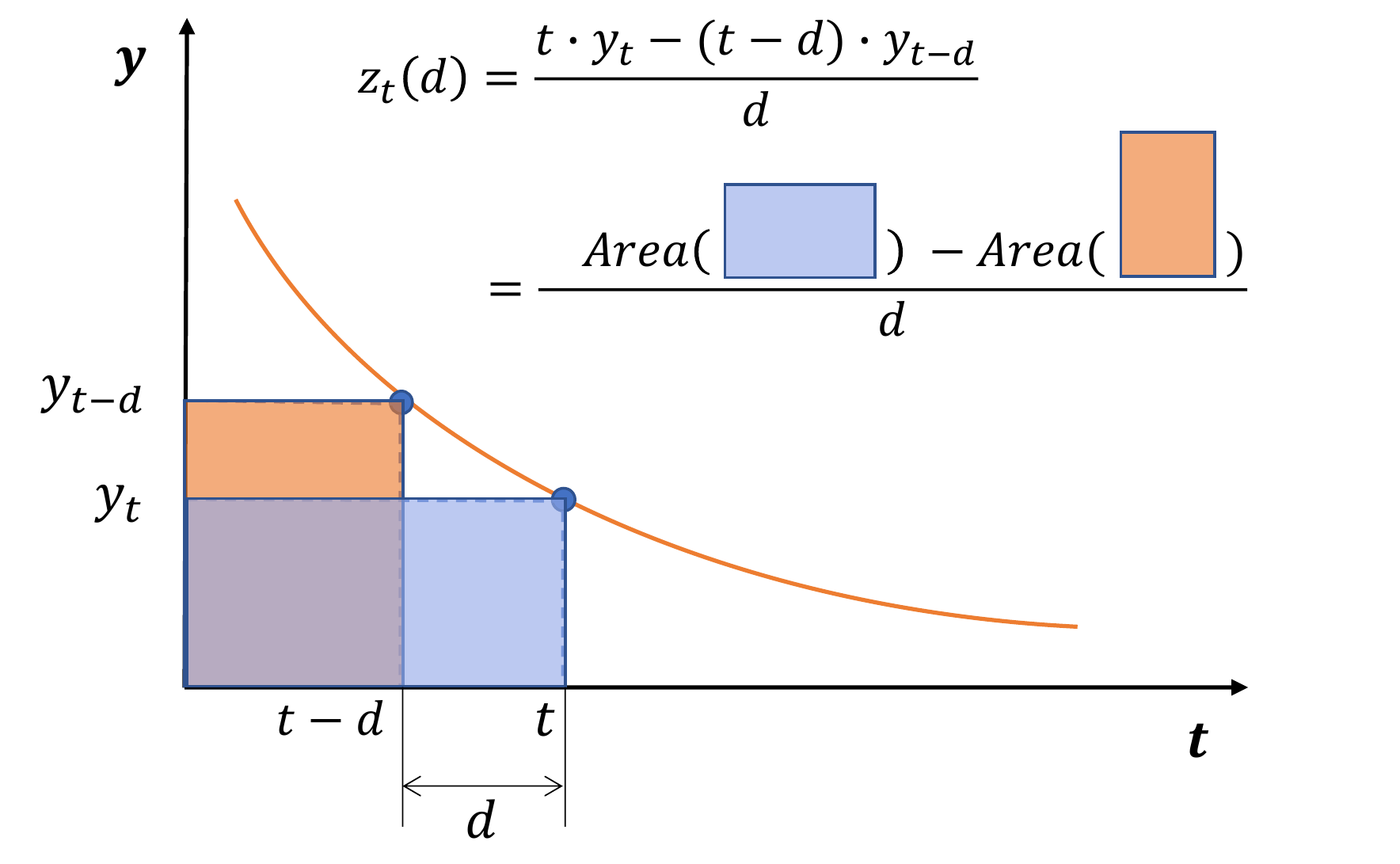}
    \caption{Curve transformation between LNE and WNE at position $t$ with window size $d$.}
    \label{fig:trans}
\end{figure}

\subsubsection{Lifetime NE vs. Window NE Curve}
\textit{Lifetime NE} (LNE) is evaluated over all the examples it has ever seen from the beginning and thus reflects the average model performance over the whole training process until now. 
More concretely, at the point where in total $t$ examples are used for training, the lifetime NE curve value
\begin{equation}\label{eq:LNE}
    \text{LNE}_t = \frac{\frac{1}{t}\sum_{i=1}^{t}l(q_i, \hat{q}_i)}{l(\Bar{q}_{[1,t]}, \Bar{q}_{[1,t]})}
\end{equation}
Where $\Bar{q}_{[1, t]}$ denotes the average empirical CTR of all the $t$ trained examples, and $l$ is defined in Eq. (\ref{eq:loss}).

In contrast to LNE that evaluates the average model performance throughout the whole training process, \textit{window NE} (WNE) is used to evaluate the latest model performance. It is defined as the average log loss divided by the average empirical CTR over the most recent $d$ examples where $d$ is the window size. Formally,
\begin{equation}\label{eq:WNE}
    \text{WNE}_t(d) = \frac{\frac{1}{d}\sum_{i=t - d + 1}^{t}l(q_i, \hat{q}_i)}{l(\Bar{q}_{[t - d + 1,t]}, \Bar{q}_{[t - d + 1,t]})}
\end{equation}
where $\Bar{q}_{[t-d+1, t]}$ denotes the average empirical CTR of the most recent $d$ trained examples.
By comparing Eq. (\ref{eq:LNE}) and (\ref{eq:WNE}), we can see that LNE can be view as a special WNE where the window size $d=t$. 

\section{Proposed Approach}

\begin{figure}[t]
    \centering
    \includegraphics[width=0.95\linewidth]{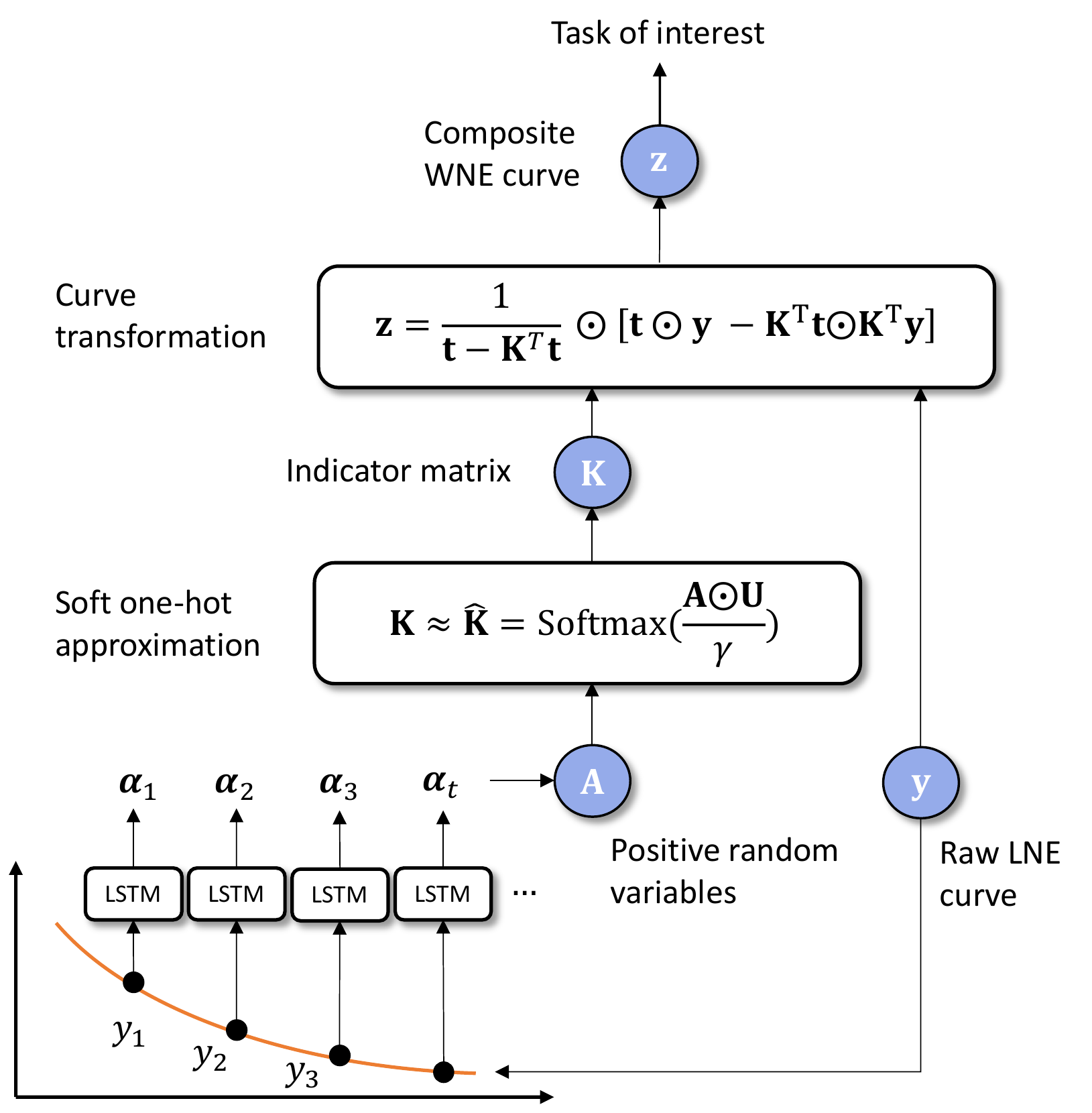}
    \caption{Overview of \ours layer.}
    \label{fig:trans_layer}
\end{figure}

In this section, we introduce the proposed \ourmodels framework. We first illustrate the motivation of proposing the adaptive curve transformation (ACT) approach in Section \ref{sec:motiv}. Then we introduce the ACT layer and its optimization in Section \ref{sec:ACT} to \ref{sec:individualize}. We finally introduce the proposed difference-based relative curve ranking architecture.
Figure \ref{fig:trans_layer} shows the overview of the proposed ACT layer, and Figure \ref{fig:diff} illustrates the components of the difference-based curve ranker.

\subsection{Motivation}\label{sec:motiv}
The WNE values for each curve point can be derived from the observed LNE curve. Proposition \ref{prop:ne} explains the relationship between LNE and WNE. The proof can be found in the appendix.

\begin{proposition}[Relationship between LNE and WNE]\label{prop:ne}
    Assume the average empirical CTR remain stable along the training process, i.e., $\Bar{q}[1,i] = \Bar{q}[1,j] = \Bar{q}[i+1,j] (i < j)$, the WNE and LNE follow the following transformation between each other:
    \begin{equation}
        \text{WNE}_t(d) = \frac{t\cdot\text{LNE}_t - (t-d)\cdot\text{LNE}_{t-d}}{d}
    \end{equation}
    where $d$ is the window size and $\text{LNE}_{t-d}$ denotes the lifetime NE value at point $t-d$.
\end{proposition}

In addition to leveraging the observed LNE curve, we propose incorporating the corresponding WNE values into our ranking model for two reasons.
Firstly, LNE curves are typically observed as a global perspective of the training dynamics across the entire training process, while WNE provides a closer look at a short period of training time and is better at characterizing the latest local model performance. The discrepancy between input curves (LNE) and target ranking results (associated with WNE) can lead to suboptimal ranking results. 
Secondly, both LNE and WNE with a certain window size can contain predictive information about the final model performance over different ranges of time. To have a comprehensive understanding of the training performance, both should be considered. 

To incorporate WNE into the ranking model, a natural way is to directly use a transformed WNE curve with a certain window size as the model input. However, it completely ignores the potential predictive power of LNE, which contains the global patterns of the curve. Most importantly, using a uniform window size for all curves and the whole training process leads to suboptimal results, because for different curves or different positions at the same curve, the dynamics and optimal window sizes are different. 
To address this issue, we propose using an adaptive approach to optimize the window size $d$ as a trainable parameter alongside the training of the ranker model for better ranking performance. In the following sections, we introduce our proposed \underline{A}daptive \underline{C}urve \underline{T}ransformation (\ours) framework. 

\subsection{Adaptive Curve Transformation}\label{sec:ACT}
We first relax the notion of the optimal window size to be position-specific, meaning that at a different point of model training, the corresponding best window size may be different. Given a partially observed LNE learning curve, $\mathbf{y} = [y_1, y_2, ..., y_L]^T$, consider the corresponding window sizes for each curve point are $\mathbf{d} = [d_1, d_2, ..., d_L]^T$ where $d_t \leq t$. Here $\mathbf{d}$ is considered as a trainable parameter. It's worth noting that when $d_t=t$, WNE is equivalent to LNE. 
Based on Proposition \ref{prop:ne}, the corresponding WNE values $\mathbf{z} = [z_1, z_2, ..., z_L]^T$ can be calculated via the following window transformation
\begin{equation}\label{eq:trans}
    \mathbf{z} = \frac{1}{\mathbf{d}}\odot \left[\mathbf{t}\odot \mathbf{y} - (\mathbf{t}-\mathbf{d})\odot \text{IndexSelect}(\mathbf{y}, \mathbf{t} - \mathbf{d})\right]
\end{equation}
where $\mathbf{t} = [1, 2, ..., L]^T\in\mathbb{R}^L$ are the ordered indices, $\odot$ denotes element-wise multiplication, and 
\begin{equation}\label{eq:indexing}
    \text{IndexSelect}(\mathbf{y}, \mathbf{k}) = [y_{k_1}, y_{k_2}, ..., y_{k_L}]^T\in\mathbb{R}^L
\end{equation}
is the indexing function that selects elements from $\mathbf{y}$ at indices $\mathbf{k} = [k_1, k_2, ..., k_L]^T$. The curve transformation at each point in Eq. (\ref{eq:trans}) can be depicted as Figure \ref{fig:trans}.

However, the curve transformation layer in Eq. (\ref{eq:trans}) struggles to flexibly update the window variables $\mathbf{d}$ to optimal values during neural backpropagation. Because the window sizes $\mathbf{d}$ are discrete variables, and the indexing function $\text{IndexSelect}(\mathbf{y}, \mathbf{k})$ is non-differentiable w.r.t. indices $\mathbf{k}$. Gradients can not flow freely through the window transformation layer defined in Eq. (\ref{eq:trans}) when training the ranking model.

\subsection{Differentiable Soft One-hot Indexing}\label{sec:indexing}

To update the list of windows $\mathbf{d}$ flexibly and smoothly during model training via gradient descent based optimization process, we propose a differentiable continuous relaxation for the discrete indexing function $\text{IndexSelect}(\mathbf{y}, \mathbf{k})$ via a reparameterization trick. This relaxation allows a low-variance estimate of the gradient through discrete stochastic indices by reparameterizing the selection of indices into the update of a continuous random variable matrix and then differentiating through those random variables.

Let's define $\mathbf{e}_{k}\in\mathbb{R}^L$ as the one-hot vector with the $k$-th entry being 1, e.g., $\mathbf{e}_3 = [0, 0, 1, 0, ..., 0]^T\in\mathbb{R}^L$. We can use this one-hot vector to select the $k$-th entry from an arbitrary vector $\mathbf{s} = [s_1, s_2, ..., s_L]\in\mathbb{R}^L$ by the inner product $s_k = \mathbf{e}_k^T\mathbf{s}$.
Now, we rewrite the window sizes $\mathbf{d}$ using their left indices for each window, $\mathbf{d} = \mathbf{t} - \mathbf{k}$, where the left indices $\mathbf{k} = [k_1, k_2, ..., k_L]^T$. 
Using the one-hot vectors defined above, it's straightforward to represent the left indices as
\begin{equation}\label{eq:k}
    \mathbf{k} = \left[\mathbf{e}_{k_1}, \mathbf{e}_{k_2}, ..., \mathbf{e}_{k_L}\right]^T\mathbf{t} = \mathbf{K}^T\mathbf{t} 
\end{equation}
where $\mathbf{K}=\left[\mathbf{e}_{k_1}, \mathbf{e}_{k_2}, ..., \mathbf{e}_{k_L}\right]\in\mathbb{R}^{L\times L}$ is the corresponding \textit{indicator matrix} with $\mathbf{e}_{k_t}$ as the $t$-th column one-hot vector. 
Similarly, we can rewrite the indexing function in Eq. (\ref{eq:indexing}) as
\begin{equation}\label{eq:indselect}
    \text{IndexSelect}(\mathbf{y}, \mathbf{k}) = \mathbf{K}^T\mathbf{y}
\end{equation}
With Eq. (\ref{eq:k}) and (\ref{eq:indselect}), we can rewrite the window transformation layer in Eq. (\ref{eq:trans}) to
\begin{equation}\label{eq:new_trans}
    \mathbf{z} = \frac{1}{\mathbf{t} - \mathbf{K}^T\mathbf{t}}\odot \left[\mathbf{t}\odot \mathbf{y} - (\mathbf{K}^T\mathbf{t}) \odot (\mathbf{K}^T\mathbf{y})\right]
\end{equation}

The idea behind the differentiable continuous relaxation is to relax the one-hot vector $\mathbf{e}_k$ with binary entries into a continuous random probability vector $\hat{\mathbf{e}}_k$ where all entries are between 0 and 1 and sum to 1, e.g., $\hat{\mathbf{e}}_3 = [0.001, 0.002, 0.981, 0.003, ..., 0.001]^T\in\mathbb{R}^L$. A specifically modified $\text{Softmax}$ function can enable this relaxation. 
Given positive random matrix $\mathbf{A} = [\boldsymbol{\alpha}_1, \boldsymbol{\alpha}_2, ..., \boldsymbol{\alpha}_L] \in\mathbb{R}^{L\times L}$ where $\boldsymbol{\alpha}_t \in (0, 1)^L$, we can use the following relaxation to smoothly approximate the true indicator matrix $\mathbf{K}$
\begin{equation}\label{eq:softmax}
    \mathbf{K} \approx \hat{\mathbf{K}} = \text{Softmax}(\frac{\mathbf{A}\odot\mathbf{U}}{\gamma})
\end{equation}
where $\text{Softmax}(\cdot)$ operates Softmax computation in each column of the input matrix,
and $\mathbf{U}\in\mathbb{R}^{L\times L}$ is an upper triangular mask matrix (having 0s under the main diagonal) to help make sure the left index keeps on the left of the current position, $1 \leq k_t \leq t$. The upper triangle matrix $\mathbf{U}$ has entries
\begin{equation}
    u_{ij} = 
    \begin{cases}
      1 & \text{if\ } i \leq j\\
      0 & \text{if\ } i > j
    \end{cases}
\end{equation}
The hyperparameter $\gamma\in(0, 1]$ is the \textit{smoothness factor} to control the degree of relaxation. 
In Eq. (\ref{eq:softmax}), the \textit{soft indicator matrix} $\hat{\mathbf{K}}$ smoothly approaches the discrete true indicator matrix $\mathbf{K}$ as $\gamma \rightarrow 0$. Increasing $\gamma$ will create "smoother" probability distributions that are less concentrated around the positions of the largest input values in each $\hat{e}_k$.
When $\gamma = 1$, Eq. (\ref{eq:softmax}) reduces to standard Softmax. We select $\gamma$ from $\{0.005, 0.01, 0.05, 0.1\}$ and find that $\gamma = 0.05$ works the best for our dataset.

In the random variable matrix $\mathbf{A}$, the relative numerical order of the entries in the column vector $\boldsymbol{\alpha}_t$ controls the output one-hot $\hat{\mathbf{e}}_{k_t}$ and therefore the corresponding index $k_t$.
The softmax-based approximation amplifies the difference between the entries of each column vector in $\mathbf{A}$ while preserving the relative numerical order of the masked variables $\mathbf{A}\odot\mathbf{U}$.
A relatively small value change in the random variables $\boldsymbol{\alpha}_t$ can result in a more drastic change of the soft one-hot $\hat{\mathbf{e}}_{k_t}$ and therefore a clear shift of the index $k_t$.
Thus, by reparameterizing the stochastic selection of indices using the random variables $\mathbf{A}$, gradients are allowed to flow through the ACT layer.
With Eq. (\ref{eq:new_trans}) and (\ref{eq:softmax}), the optimization of the windows $\mathbf{d}$ is converted into the problem of optimizing the positive random variables $\mathbf{A} \in (0, 1)^{L\times L}$ that controls the left indices of each window. This differentiable ACT layer can be optimized jointly with any downstream ranking model during model training.

\subsection{Individualized Windows with Seq2seq Parameterization}\label{sec:individualize}
Instead of explicitly optimizing one single random matrix $\mathbf{A}=[\boldsymbol{\alpha}_1, \boldsymbol{\alpha}_2, ..., \boldsymbol{\alpha}_L]$ that generates the same set of position-specific windows for all curves, we can further enhance the flexibility of ACT to be also curve-specific. It can be achieved by conditioning the window size $d_t$ on the observed input curve points using a neural network (NN)
\begin{equation}
    \boldsymbol{\alpha}_t = \text{NN}_{\boldsymbol{\phi}}([y_1, y_2, ..., y_t]^T)
\end{equation}
where $\boldsymbol{\phi}$ denotes the network parameters. The optimization of window sizes $\mathbf{d}$ is realized through optimizing the network parameters $\boldsymbol{\phi}$. 
To accommodate the consideration that the generated variable at the $t$-th point, $\boldsymbol{\alpha}_t\in\mathbb{R}^L$, is only associated with the current and previous curve points, a natural design choice is to use a sequence-to-sequence (seq2seq) model, such as LSTM, to instantiate this NN. The hidden state of LSTM at each step is used to represent $\boldsymbol{\alpha}_t$
\begin{equation}\label{eq:lstm}
\begin{split}
\mathbf{h}_t &= \text{LSTM}(y_t, \mathbf{h}_{t-1}; \boldsymbol{\phi})\\
\boldsymbol{\alpha}_t &= \frac{e^{\mathbf{h}_t}}{1 + e^{\mathbf{h}_t}}
\end{split}
\end{equation}
where $\boldsymbol{\phi}$ denotes the trainable parameters of the LSTM, the sigmoid function ensures $\boldsymbol{\alpha}_t \in (0, 1)^L$. With the seq2seq parameterization in Eq. (\ref{eq:lstm}), each curve can have specific windows for transformation optimized w.r.t. its unique curve dynamics and temporality.

Figure \ref{fig:trans_layer} illustrates the overview of the proposed ACT layer for use in ranking models. The ACT layer can be easily integrated with any curve ranking model, such as the LCRankNet model~\cite{wistuba2020learning}, by replacing the original curves with the transformed ones. All the parameters of ACT can be optimized alongside those of the curve ranker during training to improve ranking performance.

\subsection{Difference-based Relative Curve Ranking}
In Siamese-based ranking methods, such as LCRankNet~\cite{wistuba2020learning}, each twin neural network in the structure serves as a scoring function, and the pairwise ranking result is based on the difference between the output scores of the two networks. The ranking effectiveness is therefore highly dependent on the accuracy of the scoring function in predicting network performance.
These approaches may struggle to accurately rank curves with small differences, especially if the differences are not consistent over the entire curve. This is because the scoring model may not be able to capture subtle relative patterns in the data.
In addition, the scoring function leverages the full dynamics and temporal patterns of the learning curves to predict network performance, but not all of these patterns necessarily contribute to the final superiority of curves. 
In contrast, the dynamics and temporality of the difference are more important compared to those of each individual curve in predicting final superiority, because the relative performance is what ultimately determines which curve is better. 

\begin{figure}[t]
    \centering
    \includegraphics[width=0.99\linewidth]{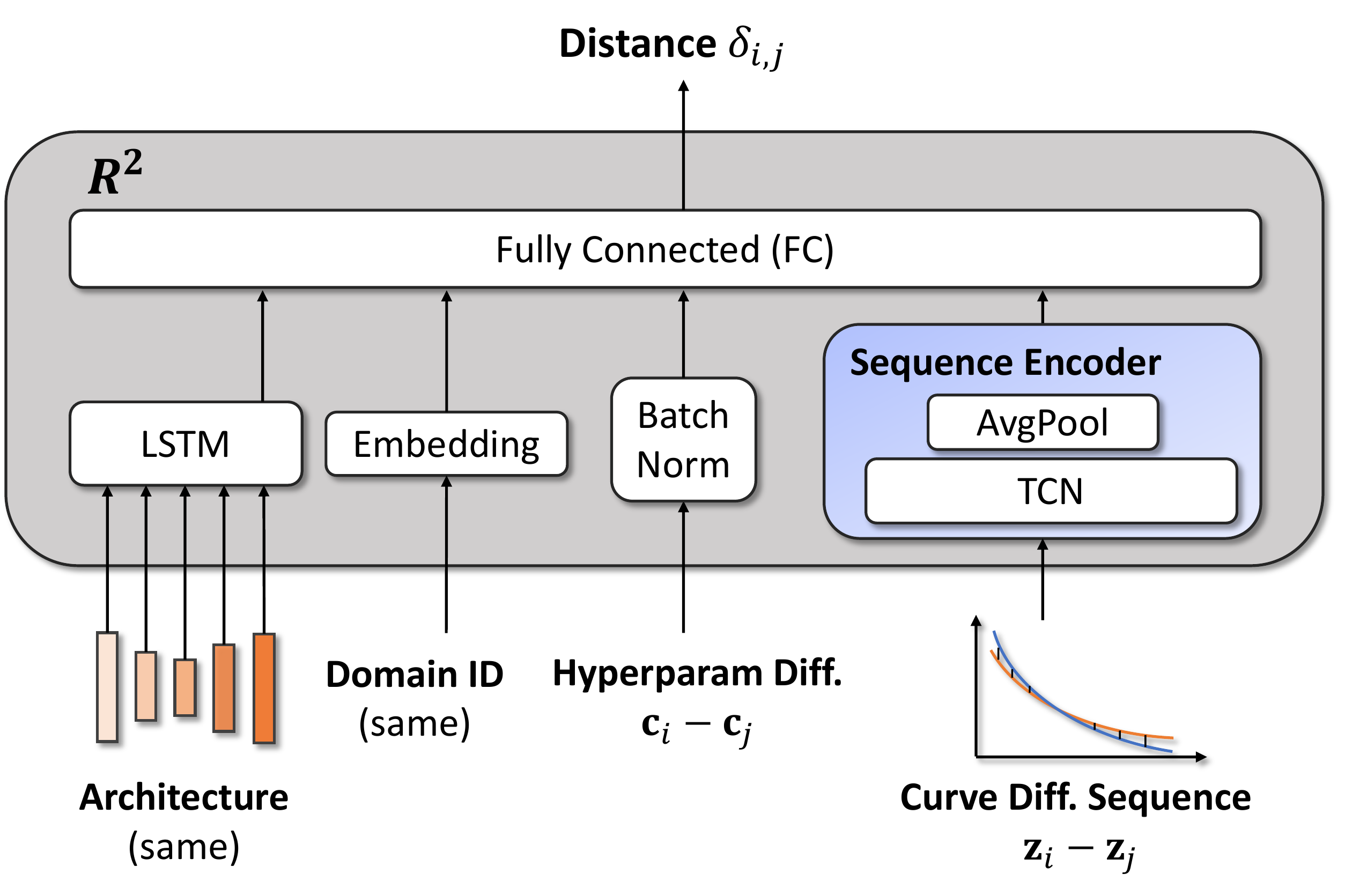}
    \caption{Architecture of \diffranker.}
    \label{fig:diff}
\end{figure}

Based on the above observations, we propose directly using the difference between the two raw learning curves as the basis of calculating the distance $\delta$ for pairwise ranking.
We refer to this difference-based model as \underline{R}elative curve \underline{R}anking (\diffranker).
Our proposed \diffranker can be more effective in situations where the difference between the two curves is small and inconsistent, as it allows the ranking model to focus on the relative performance of the two curves over time, so that the model can better capture the changes in relative performance that may not be evident in the individual curves themselves. The overview of \diffranker is shown in Figure \ref{fig:diff}. It contains four components:
\begin{itemize}
    \item \textbf{Learning curve}: The differences of the two curves are directly used as input. These differences are fed into a Temporal Convolutional Network (TCN)~\cite{bai2018empirical} as the curve encoder, with an average pooling layer at the end for feature aggregation.
    
    \item \textbf{Hyperparameters}: The differences of hyperparameters (e.g., learning rate) between the two networks are fed into a batch normalization~\cite{ioffe2015batch} layer.
    
    \item \textbf{Architecture}: We use an LSTM to aggregate the architecture information~\cite{Zoph_2018_CVPR,zoph2017neural}. The characteristics (e.g., dimensions) of each layer are organized into a sequence and fed into the LSTM. The last hidden state of the LSTM is used as the architecture embedding. Note that in our dataset (see Section \ref{sec:data}), the autotuning jobs only tune the hyperparameters of the model, so the architecture of the two curves being compared is identical. As a result, the architecture embedding is only used once in this case. For the cases where the architecture is different, we recommend using the difference of the two architecture embeddings instead.
    
    \item \textbf{Domain ID}: In order to address curves across domains (e.g., model type, dataset, application),
    the domain IDs are passed to an embedding layer that maps each ID to a specific vector representation, which can be optimized through the training process. This helps model the domain-specific peculiarities in order to adjust the ranking when necessary.
    When the domain is new (i.e., not existing during training), the domain embedding is initialized at random.
    In this work, we use network type as the domain ID since we train the ranking model across different model types.
\end{itemize}
The outputs of the above four network components are then concatenated and passed to a fully connected (FC) layer. 
Compared to LCRankNet, we replace the calculation of distance $\delta_{i,j}$ in Eq. (\ref{eq:delta}) with the direct output
\begin{equation}
    \delta_{i,j} = h((\mathbf{x}_i, \mathbf{y}_i), (\mathbf{x}_j, \mathbf{y}_j))
\end{equation}
where $h(\cdot)$ is the distance function used by \diffranker shown in Figure \ref{fig:diff} that directly evaluates the distance between $m_i$ and $m_j$.
Then the probability of final superiority is calculated using Eq. (\ref{eq:prob}) and the parameters are optimized via minimizing the cross-entropy loss in Eq. (\ref{eq:ce_loss}).
During training, the architecture embedding is further passed to an LSTM decoder for architecture reconstruction in order to stabilize the training as in~\cite{wistuba2020learning}. The overall loss is
\begin{equation}
    \mathcal{L} = \mathcal{L}_{\text{CE}} + \lambda\mathcal{L}_{\text{MSE}}
\end{equation}
where $\mathcal{L}_{\text{MSE}}$ denotes the mean square error (MSE) loss of the architecture reconstruction, and trade-off hyperparameter $\lambda$ is used to control the involvement of each component. We select the value of $\lambda$ from $\{10^{-4}, 10^{-3}, ..., 10^{4}\}$, and find that $\lambda = 10^{-1}$ works the best for our dataset.

\section{Experiment}

\subsection{Data}\label{sec:data}
\subsubsection{Data Collection}
We collected data on NE learning curves and model configurations from the internal model autotuning platform at Meta. This platform uses Bayesian optimization~\cite{snoek2012practical} to tune the hyperparameters of various deep learning models used for online advertising and recommendation on social media platforms such as Facebook and Instagram.
We collected data including the following four components:
\begin{itemize}
    \item \textbf{Learning curve}: During model training, LNE of the current model on the new batch of training examples is calculated at each time. The learning curve records a list of LNE values versus the corresponding number of examples.
    \item \textbf{Hyperparameter}: Model hyperparameters, especially optimizer parameters (e.g., learning rate, epsilon, decay rate).
    \item \textbf{Architecture}: Architecture information including dimensionality of embedding at each component of networks.
    \item \textbf{Characteristics}: Other categorization features, such as network type and autotuning job IDs.
\end{itemize}
In this work, we collected data from in total 4,771 networks with different model configurations, which correspond to over 74 thousand curve pairs. The overall statistics of the collected dataset is shown in Table \ref{tab:dataset} .

\begin{table}[t]
    \centering
    \fontsize{8.5pt}{10pt}\selectfont
    \caption{Overall statistics of dataset.}
    \begin{tabular}{||ll||}
    \hline
    \textsc{Statistics} & \textsc{Number}\\
    \hline\hline
    \# networks (curves) & 4,771 \\
    \# network (curve) pairs & 74,856  \\
    \# autotuning jobs & 79 \\
    \# types of networks & 28 \\
    \# types of hyperparameters & 13 \\
    lengths of curves (mean $\pm$ std) & 163 $\pm$ 193 \\
    \hline
    \end{tabular}
    \label{tab:dataset}
\end{table}

\subsubsection{Curve Pairing and Labeling}
Our ranking model is designed to rank pairs of curves, which are generated by the same autotuning job. 
Therefore we group every two curves under the same autotuning job into pairs.
To train the model, we assign a binary label to each curve pair based on their final \textbf{1-day WNE} values at the end of training. 
1-day WNE is the de facto criterion widely used in tuning online advertising models, which uses the number of examples of one day data as the window size.

\subsubsection{Data Preprocessing and Experiment Setup}
To ensure the reliability of our model, we only include curves with at least 10 observed measurements throughout the training process in our analysis. To simplify the data, we resample all curves to have 100 points.
We use 5-fold cross validation for performance evaluation. In each fold, the curves and their corresponding configuration data are grouped by autotuning jobs and then split into training, validation, and test sets in a 70:10:20 ratio, ensuring that no curves from the same autotuning job are split into different subsets. This allows us to estimate the out-of-sample performance of the trained ranker model on a new autotuning task.
To evaluate the effectiveness of the model, we used both the percentage of correctly classified curve pairs (accuracy) and the area under the receiver operating characteristic curve (AUC). We report the mean and standard deviation of the 5-fold cross-validation results for all models.

\subsection{Data Characteristics and Naive Baseline}
\begin{figure}[t]
    \centering
    \includegraphics[width=.8\linewidth]{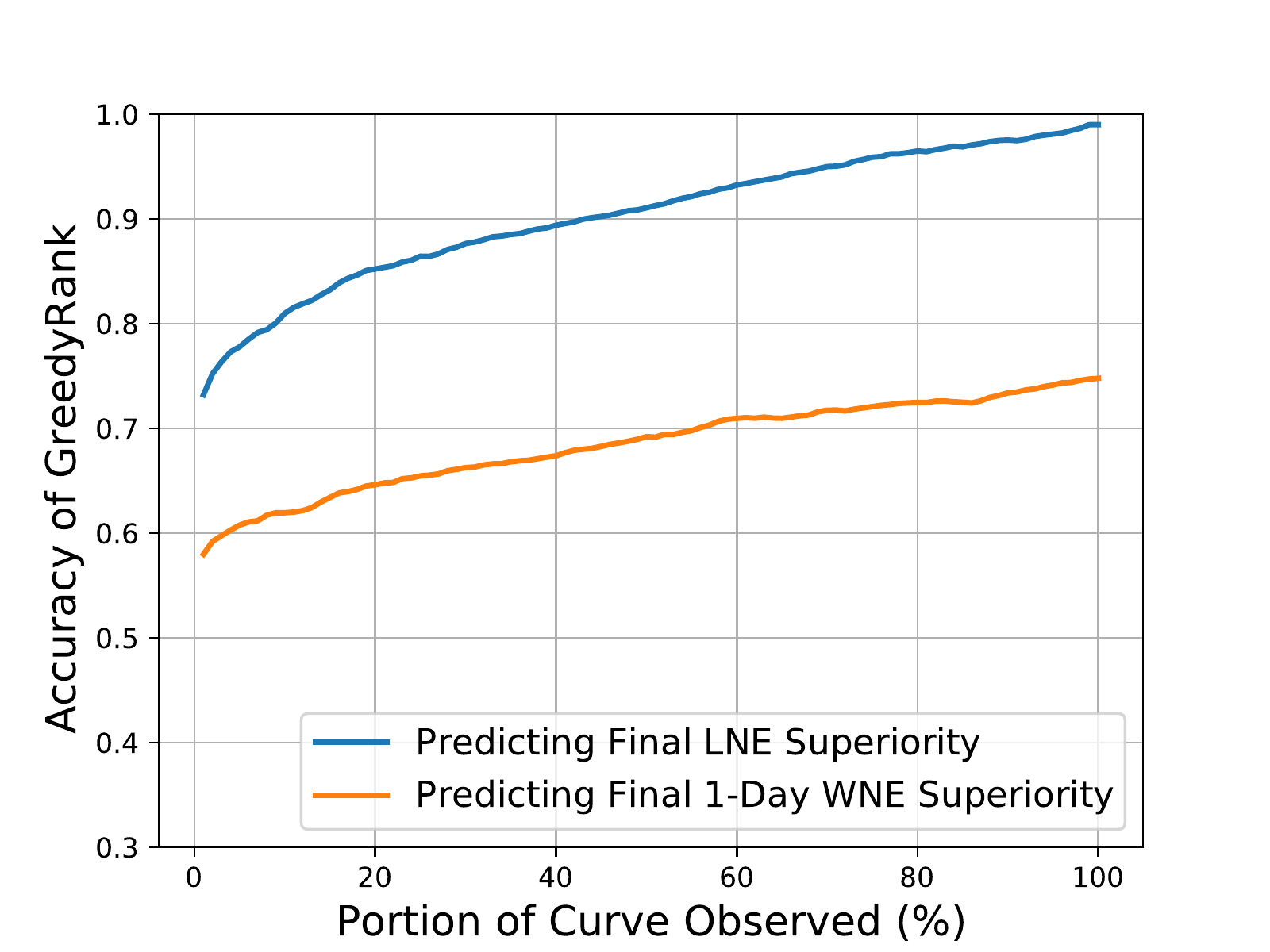}
    \caption{Accuracy of GreedyRank, i.e., portion of curve pairs with current relative superiority being consistent with ground truth (LNE or 1-day WNE is used to determine ground-truth superiority).}
    \label{fig:naive}
\end{figure}

Previous work~\cite{klein2016learning} noted that
using the last seen value to determine the ranking of learning curves is a simple yet efficient method, which is similar to the approach often used by human experts to determine when to stop training a model.
We refer to this greedy method as GreedyRank. 
It's worth noting that the last observed value is a very good predictor in general, since the superiority of most curve pairs remain consistent over time.
After all, it is the only one used by automated optimization methods like  Hyperband~\cite{li2017hyperband} and Successive Halving~\cite{jamieson2016non}.
It ranks pairs incorrectly when the superiority of the two curves changes at a future point.
Figure \ref{fig:naive} shows the characteristics of the collected curve dataset by displaying the baseline accuracy of GreedyRank under different definitions of curve superiority.
It also shows that the ranking accuracy increases when longer curves are observed for both cases, since the superiority of the curves tends to stabilize over time.

While this simple strategy is effective for most curve pairs where the dominance of one curve is consistently apparent, we argue that it may not be as effective in our NE curve data, especially when WNE is used as the criterion for final model performance.
As shown in Figure \ref{fig:naive}, the accuracy of GreedyRank when 1-day WNE is used as the labeling criterion is consistently lower than when LNE is used.
For example, a model that shows better updated performance on recent examples (better WNE) but has worse performance at the beginning may still have overall worse average performance (worse LNE) throughout the entire curve.
The low accuracy of 0.75 even when 100\% curves are observed reflects the fact that 25\% of the curves end up with inconsistent LNE values and 1-day WNE values.
The discrepancy between the observation and outcome will lead to a higher ranking error rate for GreedyRank and potentially other existing ranking approaches that do not consider the unique characteristics of NE curves, such as LCRankNet~\cite{wistuba2020learning}.

\subsection{Ranking Performance}
We compare our proposed \ourmodels framework and its variants to GreedyRank~\cite{klein2016learning} and the state-of-the-art LCRankNet~\cite{wistuba2020learning} on ranking model configurations from new autotuning tasks. We did not include pointwise methods~\cite{chandrashekaran2017speeding,domhan2015speeding,klein2016learning,baker2017accelerating} in the comparison because previous research~\cite{wistuba2020learning} has shown they perform significantly worse than LCRankNet.

\subsubsection{Overall Performance}
Table \ref{tab:main} summarizes the ranking performance of all the models. Our proposed \ourmodels achieves \textbf{13\% to 15\% improved average accuracy} over GreedyRank given different lengths of observed learning curves from new autotuning jobs. Compared to LCRankNet, our proposed \ourmodels consistently achieves \textbf{around 5\% improved accuracy and around 7\% improved AUC} on average. The significant improvement over the baseline methods shows the overall effectiveness of our proposed ACT mechanism and \diffranker ranking architecture.

\begin{table*}[t]
  \caption{Cross-validation results (mean $\pm$ std) of pairwise ranking across autotuning jobs.}
\fontsize{8.5pt}{11pt}\selectfont
\begin{tabular}{||l|cc|cc|cc||}
\hline
 \% \textsc{Curve Observed} &
  \multicolumn{2}{c|}{20\%} &
  \multicolumn{2}{c|}{40\%} &
  \multicolumn{2}{c||}{60\%} \\
 \textsc{Metric} &
  \multicolumn{1}{c}{AUC} &
  \multicolumn{1}{c|}{\textsc{Accuracy}} &
  \multicolumn{1}{c}{AUC} &
  \multicolumn{1}{c|}{\textsc{Accuracy}} &
  \multicolumn{1}{c}{AUC} &
  \multicolumn{1}{c||}{\textsc{Accuracy}} \\
\hline\hline
GreedyRank~\cite{klein2016learning}       &  N.A.  & .6461 (.0529) & N.A & .6738 (.0455) & N.A. & .7096 (.0460) \\
LCRankNet~\cite{wistuba2020learning} w/o Metadata & .6930 (.0664) & .6636 (.0460) & .8267 (.0202) & .7599 (.0140) & .8631 (.0216) & .7971 (.0189) \\
LCRankNet~\cite{wistuba2020learning}              & .7821 (.1091) & .7289 (.0762) & .8308 (.0261)
 & .7607 (.0165)  & .8793 (.0276) & .8067 (.0205) \\
\hline\hline
LCRankNet-ACT            & .7978 (.0469) & .7235 (.0391) & .8640 (.0237) & .7825 (.0201)  & .8978 (.0281) & \underline {.8262 (.0267)} \\
\diffranker w/o Metadata & .8061 (.0195) & .7267 (.0142) & .8584 (.0099) & .7791 (.0118) & .8899 (.0414) & .8131 (.0354) \\
\diffranker (\ourmodels w/o ACT)              & \underline {.8352 (.0258)} & \underline {.7525 (.0221)} & \underline {.8693 (.0117)} & \underline {.7878 (.0103)}  & \underline {.9057 (.0263)} & .8210 (.0292) \\
\ourmodels             & \textbf {.8546 (.0303)} & \textbf {.7706 (.0256)} & \textbf {.9013 (.0136)} & \textbf {.8158 (.0123)}  & \textbf {.9307 (.0254)} & \textbf {.8538 (.0277)} \\
\hline
\end{tabular}
\label{tab:main}
\end{table*}

\subsubsection{Adaptive Curves vs. Raw Curves}
In Table \ref{tab:main}, LCRankNet-ACT denotes the method that combines our ACT layer with LCRankNet, using the transformed composite WNE curves as input. The parameters of both LCRankNet and ACT are trained jointly. On the other hand, \ourmodels w/o ACT denotes the version without the ACT layer, which uses the raw LNE curves as input of our proposed difference-based ranking model (\diffranker). The results show that LCRankNet-ACT consistently outperforms the original LCRankNet across different curve lengths with 1\%-2\% improved AUC scores, and \ourmodels exceeds \diffranker by around 2\%-3\% improved AUC scores. This demonstrates that our proposed ACT layer can effectively improve the ranking performance by transforming the curves in a way that adapts seamlessly to the ranking model and makes it easier to differentiate between two curves. Notably, the performance improvement is more pronounced when using ACT with \diffranker. 
The improvement by using ACT on either ranking model is in general more significant when longer curves are observed. This is because the curve transformation is less flexible for the points near the curve start. Bounded by the upper triangle mask $\mathbf{U}$ in Eq. (\ref{eq:softmax}), the curves points near the start tend to have more limited search space for viable windows, which in turn learn a transformed curve closer to the original LNE, resulting in a closer ranking performance.

\subsubsection{Difference-based Ranking vs. Siamese-based Ranking}
As Table \ref{tab:main} shows, the proposed \diffranker outperforms LCRankNet with an improvement of 3\%-5\% in AUC scores when using raw LNE curves. When using ACT to adaptively transform the curves, \ourmodels outperforms LCRankNet-ACT with an improvement of 3\%-6\% in AUC scores. A more significant improvement is observed in the comparison between the two "curve only" models, \diffranker w/o Metadata vs. LCRankNet w/o Metadata, in which all metadata components (e.g., hyperparameters, architecture) are excluded from the ranking model.
Specifically, with only 20\% learning curves observed and no metadata available, \diffranker outperforms LCRankNet with an improvement of around 14\% in AUC.
Our difference-based ranking method shows higher average performance and lower variance across model groups than the Siamese-based ranking used by existing approaches. The improvement in both average performance and the variance is more significant when shorter curves (20\%) are observed. This is because that difference between different curves are usually more pronounced at the early stage as shown in Figure~\ref{fig:curve} in Appendix, difference-based ranking can accurately capture this difference and the advantage is more noticeable on shorter curves.
This improvement suggests that our difference-based approach is more effective at accurately ranking curves using relative differences than existing Siamese-based methods.
By focusing on the relative performance of two curves rather than the absolute performance of each curve individually, the ranking model is more capable of capturing subtle patterns directly correlated with the final superiority of the curves. In addition, by only taking into account the difference of two curves, much irrelevant information is excluded from the input, which reduces the risk of overfitting of the ranking model, especially when the amount of training data is limited or moderate.

\subsubsection{Effect of Metadata on Ranking}
According to Table \ref{tab:main}, including metadata (hyperparameters, architecture and domain IDs) components improves the ranking performance. Only using learning curves results in a 1\%-9\% AUC decline for LCRankNet and 1\%-3\% AUC decline for our proposed \diffranker. \diffranker has shown higher robustness and stability than LCRankNet since Siamese-based ranking falls short in capturing reliable predictive information from learning curves. Both the two ranking architectures have demonstrated lower dependency on metadata in predicting the relative superiority when longer curves are observed, since more complicated patterns in curves can be discovered and used for the prediction.

\begin{figure}[t]
    \centering
    \includegraphics[width=.95\linewidth]{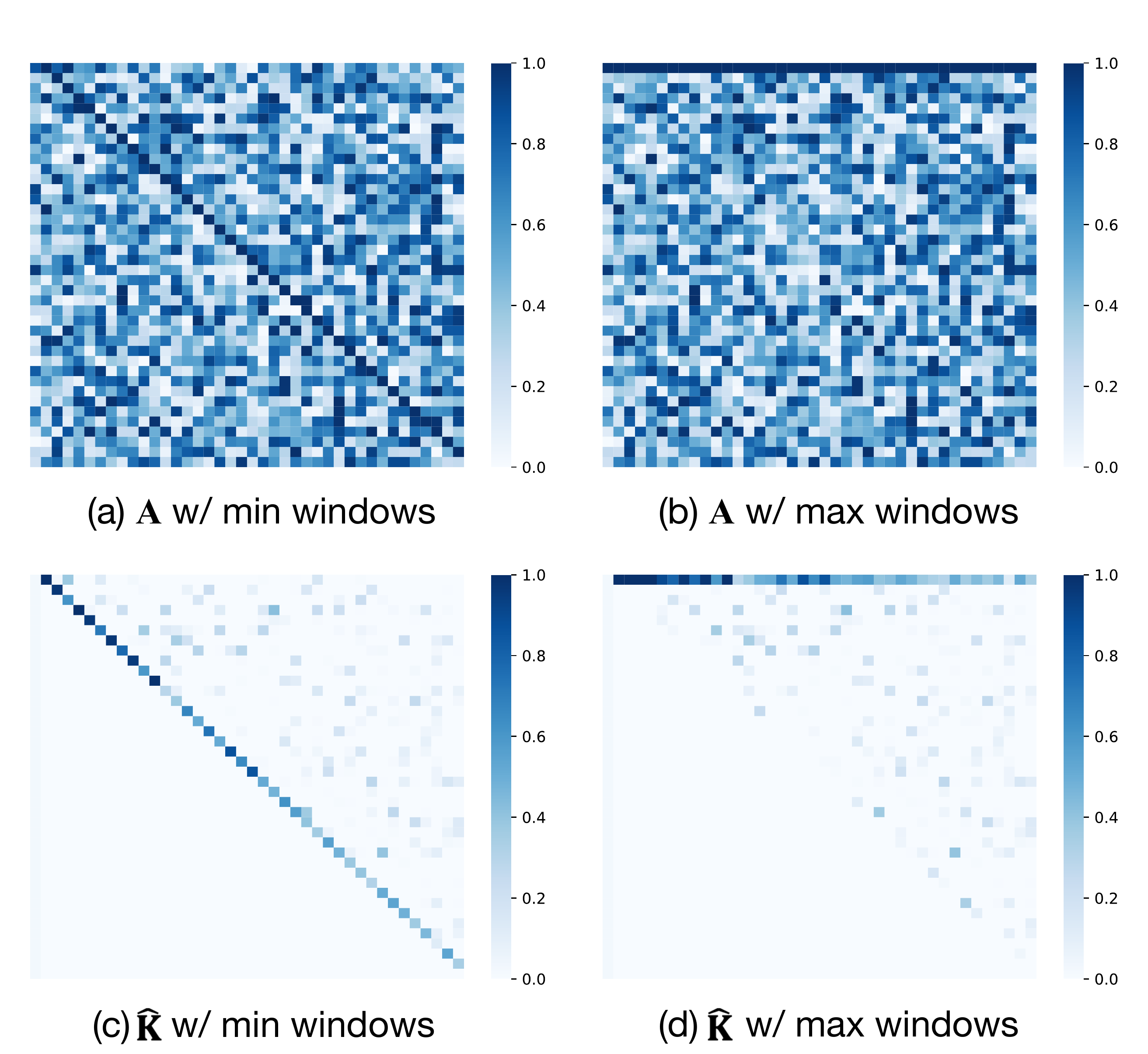}
    \caption{Random initialization of variable matrix $\mathbf{A}$ with min or max windows and the corresponding initial soft indicator matrix $\hat{\mathbf{K}}$ ($\gamma = 0.05$) for ACT-DF1. The location with the darkest color in each column of $\mathbf{A}$ and $\hat{\mathbf{K}}$ corresponds to the left-side index of the window.}
    \label{fig:heatmap}
\end{figure}

\subsection{Performance Analysis of ACT}

\begin{table*}[t]
  \caption{Comparison of ACT-based variants with different degrees of freedom for window adaptation.}
\fontsize{8.5pt}{11pt}\selectfont
\begin{tabular}{||l|cc|cc|cc||}
\hline
 \% \textsc{Curve Observed} &
  \multicolumn{2}{c|}{20\%} &
  \multicolumn{2}{c|}{40\%} &
  \multicolumn{2}{c||}{60\%} \\
 \textsc{Metric} &
  \multicolumn{1}{c}{AUC} &
  \multicolumn{1}{c|}{\textsc{Accuracy}} &
  \multicolumn{1}{c}{AUC} &
  \multicolumn{1}{c|}{\textsc{Accuracy}} &
  \multicolumn{1}{c}{AUC} &
  \multicolumn{1}{c||}{\textsc{Accuracy}} \\
\hline\hline
LCRankNet-ACT-DF1 w/ Min Init.         & .8001 (.0474) & .7256 (.0416) & \underline{.8508 (.0255)} & \underline{.7800 (.0205)} & .8834 (.0319) & .8171 (.0293) \\
LCRankNet-ACT-DF1 w/ Max Init.         & \textbf{.8126 (.0500)} & \underline{.7302 (.0428)} & .8408 (.0290) & .7745 (.0231) & \underline{.8836 (.0297)} & \underline{.8190 (.0292)} \\
LCRankNet-ACT-DF2         & \underline{.8086 (.0370)} & \textbf{.7313 (.0314)} & .8448 (.0298) & .7730 (.0234) & .8699 (.0333) & .8063 (.0237) \\
LCRankNet-ACT-DF3         & .7978 (.0469) & .7235 (.0391) & \textbf{.8640 (.0237)} & \textbf{.7825 (.0201)} & \textbf{.8978 (.0281)} & \textbf{.8262 (.0267)} \\
\hline\hline
\ourmodel-DF1 w/ Min Init.             & .8468 (.0216) & .7617 (.0204) & .8734 (.0129) & .7908 (.0114) & \underline{.9191 (.0246)} & .8346 (.0257) \\
\ourmodel-DF1 w/ Max Init.             & \underline{.8251 (.0258)} & \underline{.7438 (.0204)} & .8711 (.0123) & .7891 (.0105) & .9107 (.0271) & .8309 (.0285) \\
\ourmodel-DF2              & .8466 (.0271) & .7610 (.0188) & \underline{.8842 (.0130)} & \underline{.8005 (.0142)} & .9182 (.0243) & \underline{.8353 (.0267)} \\
\ourmodel-DF3         & \textbf{.8546 (.0303)} & \textbf{.7706 (.0256)} & \textbf{.9013 (.0136)} & \textbf{.8158 (.0123)} & \textbf{.9307 (.0254)}
 & \textbf{.8538 (.0277)} \\
\hline
\end{tabular}
\label{tab:ablation}
\end{table*}

\subsubsection{Flexibility Control of ACT}
Our proposed ACT framework (Section \ref{sec:ACT} to \ref{sec:individualize}) treats the window sizes for WNE transformation as random variables. 
We can control the flexibility of the curve transformation, i.e., \textit{degree of freedom} (DF) of learned windows, via changing how the random variables $\mathbf{A}$ in Eq. (\ref{eq:softmax}) is generated. We summarize the 3 different ACT versions with different DF of learned windows from low to high as follows:
\begin{itemize}
    \item \textbf{ACT-DF1} (curve-wise fixed, position-wise variable): Directly optimize a single $\mathbf{A}$ in Eq. (\ref{eq:softmax}) for all curves without further individualizing it to be curve-specific using neural networks as described in Section 3.4. 
    
    \item \textbf{ACT-DF2} (curve-wise variable, position-wise fixed): Individualize the generation of $\mathbf{A}$ for each curve using neural networks based on the specifics of the curve, while keeping all the column vectors in $\mathbf{A}$ the same across different curve points. 

    \item \textbf{ACT-DF3} (curve-wise variable, position-wise variable): The default setting as illustrated in Section \ref{sec:indexing} to \ref{sec:individualize}. It parameterizes $\mathbf{A}$ with neural networks for curve-specific individualization while also keeping it position-wise variable. 
\end{itemize}

We find it difficult for ACT-DF1 to explore sufficient window combinations before convergence due to the limited freedom of the variables. The learned windows tend to be highly dependent on the random initialization. To stabilize the result, we initialize the matrix $\mathbf{A}$ with random values conditional on either minimum ($d=1$, capturing most local dynamics) or maximum ($d=t$, as current curve length, equivalent to unchanged LNE curve) window sizes at the beginning of training, respectively. Figure \ref{fig:heatmap} (a) and (b) show the examples of the two ways to initialize $\mathbf{A}$. 
For ACT-DF2, it falls short in catching dynamic changes in one curve and the identical window size is unable to precisely represent different parts of a curve. The realization details of ACT-DF2 can be found in Appendix.

\subsubsection{Ablation Study of Window Flexibility}
Table \ref{tab:ablation} summarizes the ranking results of applying ACT with different window flexibility on both the Siamese-based ranking model, LCRankNet, and our proposed difference-based ranking model, \diffranker. 
In general, ACT-augmented models perform better with higher DF of windows. 
On both the two ranking architectures, ACT shows the best performance with the highest window flexibility--ACT-DF3 (the only exception is using LCRankNet with only 20\% curves observed, where less data points make training with high degree of freedom harder). 
This demonstrates that our proposed seq2seq neural parameterization (Eq. (\ref{eq:lstm})) taps into the full potential of adaptive curve transformation by enabling both the curve-wise and position-wise adaptation mechanism.
Despite the specific DF, all the variants augmented with ACT show in general better performance than the state-of-the-art method LCRankNet. This manifests the efficacy of ACT.
Particularly, the ACT layer working in conjunction with our proposed \diffranker consistently outperforms working with LCRankNet under each of the flexibility settings with a 2\%-6\% AUC improvement. This further justifies the robust effectiveness of \diffranker in various curve modifications.

\subsubsection{Effect of Soft One-hot Relaxation}
Figure \ref{fig:heatmap} visualizes the variable matrix $\mathbf{A}$ and soft indicator matrix $\hat{\mathbf{K}}$ under different window initialization. 
After the transformation by Eq. (\ref{eq:softmax}), the probability distribution of each column in $\hat{\mathbf{K}}$ is more concentrated on the position with the largest value compared to the associated column in $\mathbf{A}$. More examples of $\hat{\mathbf{K}}$ with different $\gamma$ are shown in Appendix. We can see that with smaller smoothness factor $\gamma$, the concentration effect of the probability distribution becomes more evident, and the approximation $\hat{\mathbf{K}}$ gets closer to the real indicator matrix $\mathbf{K}$.
\section{Conclusion}

In this paper, we proposed a novel learning curve ranking framework, \ourmodels, specifically tailored for ranking normalized entropy (NE) learning curves, which are commonly used in online advertising and recommendation systems. The two main novelties of the underlying model \ourmodels are the proposed differentiable self-adaptive curve transformation layer and the difference-based relative curve ranking architecture. The former allows to transform the input NE curves into composite window NE curves with the window sizes freely adapting to both different curves and different positions on the same curve, which optimizes for better ranking results. The latter directly models the difference of curves for pairwise ranking and is better at capturing subtle changes in relative performance between curves, resulting in improved performance in ranking curves with small differences.
Our extensive experiments on a real-world NE curve dataset have demonstrated the effectiveness of \ourmodels with up to 15\% improved accuracy over the widely-used greedy approach and 7\% improved AUC over the state-of-the-art method in ranking curves from new autotuning tasks.
\bibliographystyle{acm}
\bibliography{main.bib}

\clearpage

\begin{appendices}
\appendixpage

\section{Curve Examples}
\begin{figure*}[t]
    \centering
    \includegraphics[width=0.9\textwidth]{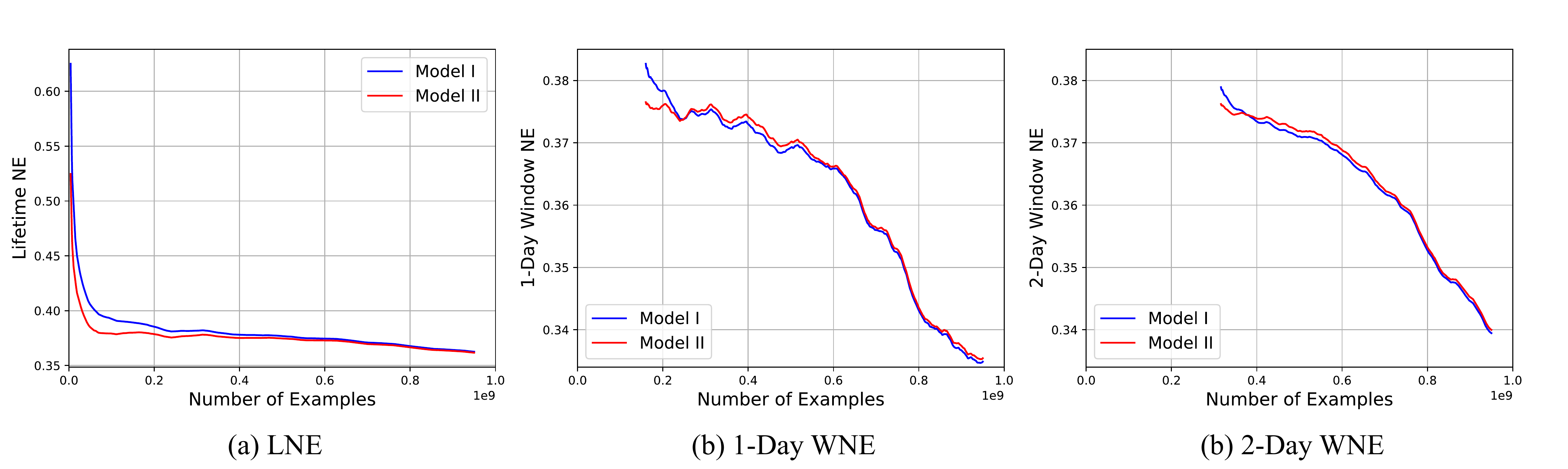}
    \caption{LNE and corresponding 1-day and 2-day WNE curves of two example model configurations.}
    \label{fig:curve}
\end{figure*}

Figure \ref{fig:curve} shows the LNE curves and the corresponding 1-day and 2-day WNE (around $0.8\times 10^9$ examples per day) curves of a typical pair of model configurations. There are no points plotted when the number of examples is less than the window size for WNE curves. We can see that the smaller the window size, the more fluctuating the curve is, and LNE curve is the smoothest since it evaluates the overall average model performance through the whole training process. Although model II has been dominating model I with lower LNE through the whole training process, the superiority of the two models in 1-day WNE has changed during the training (at around $0.25\times 10^9$ examples) and model I ended up with better 1-day WNE performance in the end. Similar superiority change was also observed on the 2-day WNE curves.

\section{Hyperparameters}
Table \ref{tab:param} lists the hyperparameters used in this paper.

\begin{table}[ht]
    \centering
    \fontsize{8pt}{10pt}\selectfont
    \caption{Hyperparameters used in \ourmodels.}
    \begin{tabular}{||l l l l l l||}
    \hline
    Batch size & 64 & Learning rate & 0.001 & Optimizer & Adam\\
    Epochs     & 100 &Dropout & 0.3 &TCN dilation & $2^k$  \\
    TCN layers & 4 & TCN filter & 64 & TCN kernal & 3\\
    LSTM layers & 2 & LSTM dim. & 64 & embed. dim. & 64\\
    $\lambda$ & 0.1 & $\gamma$ & 0.05 &  & \\
    \hline
    \end{tabular}
    \label{tab:param}
\end{table}

\section{Effect of Smoothness Factor}
\begin{figure}[th]
    \centering
    \includegraphics[width=1.\linewidth]{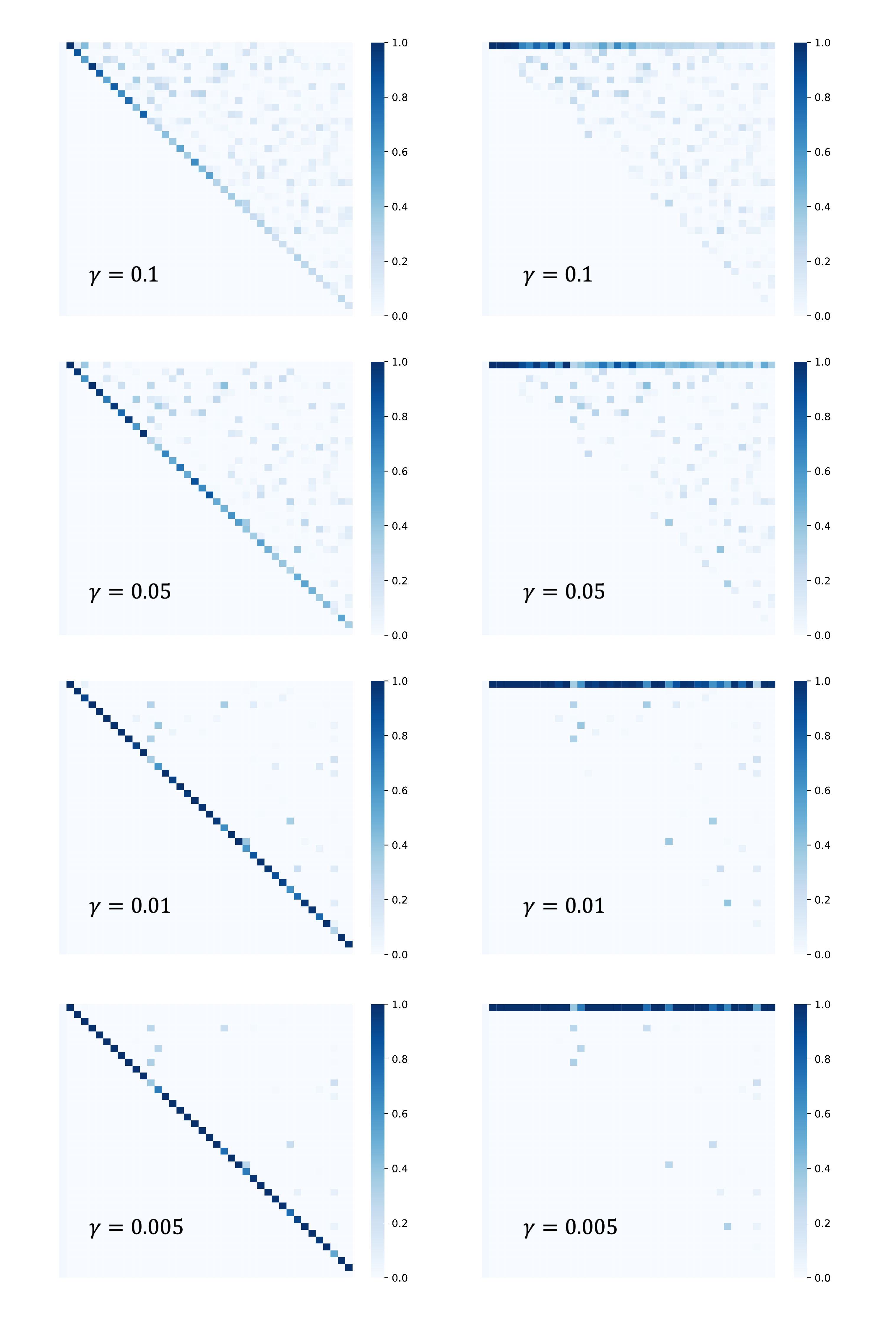}
    \caption{Soft indicator matrix $\hat{\mathbf{K}}$ generated by Eq. (\ref{eq:softmax}) with different smoothness factor $\gamma$ and random initialized $\mathbf{A}$ that corresponds to min (left column) or max (right column) windows.}
    \label{fig:heatmap_K}
\end{figure}

Examples of $\hat{\mathbf{K}}$ with different $\gamma$ are shown in Figure \ref{fig:heatmap_K}. With smaller smoothness factor $\gamma$, the concentration effect of the probability distribution becomes more evident, and the approximation $\hat{\mathbf{K}}$ gets closer to the real indicator matrix $\mathbf{K}$.

\section{Proof of Proposition 1}
Given that the average empirical CTR remain stable, we have
\begin{equation}
    \Bar{q}[1,t] = \Bar{q}[1,t-d] = \Bar{q}[t-d+1,t]
\end{equation}
Therefore,
\begin{equation}
\begin{split}
      l(\Bar{q}[1,t], \Bar{q}[1,t]) &= l(\Bar{q}[1,t-d], \Bar{q}[1,t-d])\\
      &= l(\Bar{q}[t-d+1,t], \Bar{q}[t-d+1,t])  
\end{split}
\end{equation}
According to the definition of LNE in Eq. (\ref{eq:LNE}), we have
\begin{equation}
    \sum_{i=1}^{t}l(q_i, \hat{q}_i) = t\cdot \text{LNE}_t \cdot l(\Bar{q}_{[1,t]}, \Bar{q}_{[1,t]})
\end{equation}
Thus, we can rewrite Eq. (\ref{eq:WNE}) as
\begin{equation}
\begin{split}
&\text{WNE}_t(d) = \frac{\frac{1}{d}\sum_{i=t - d + 1}^{t}l(q_i, \hat{q}_i)}{l(\Bar{q}_{[t - d + 1,t]}, \Bar{q}_{[t - d + 1,t]})} \\
&\ \ \ \ \ \ \ \ \ \ \ \ \ \ \ \ = \frac{\sum_{i=1}^{t}l(q_i, \hat{q}_i) - \sum_{i=1}^{t-d}l(q_i, \hat{q}_i)}{d\cdot l(\Bar{q}_{[t - d + 1,t]}, \Bar{q}_{[t - d + 1,t]})} \\
&= \frac{t\cdot \text{LNE}_t \cdot l(\Bar{q}_{[1,n]}, \Bar{q}_{[1,t]}) - (t-d)\cdot \text{LNE}_{t-d} \cdot l(\Bar{q}_{[1,t-d]}, \Bar{q}_{[1,t-d]})}{d\cdot l(\Bar{q}_{[t - d + 1,t]}, \Bar{q}_{[t - d + 1,t]})} \\
&= \frac{t\cdot\text{LNE}_t - (t-d)\cdot\text{LNE}_{t-d}}{d}
\end{split}
\end{equation}

\section{Realization of ACT-DF2}
For ACT-DF2, the learned windows are curve-specific but identical across different positions on the curve. We need to first be able to directly control the corresponding window sizes of the generated random variables $\mathbf{A}$ from the LSTM in Eq. (\ref{eq:lstm}). Therefore, instead of directly using $\mathbf{A}$ to generate the indicator matrix that determines the left indices of the windows as in Eq. (\ref{eq:softmax}), we can use it to directly generate the indicator matrix $\mathbf{D}$ that determines the window sizes:
\begin{equation}\label{eq:softmax_d}
    \mathbf{D} \approx \hat{\mathbf{D}} = \text{Softmax}(\frac{\mathbf{A}\odot\mathbf{U}}{\gamma})
\end{equation}
The corresponding window sizes can be calculated by $\mathbf{d} = \mathbf{D}^T \mathbf{t}$, where $\mathbf{t} = [1, 2, ..., L]^T\in\mathbb{R}^L$ are the ordered indices.
To make sure the learned windows at each position have the same size, we need the matrices $\mathbf{D}$ and $\mathbf{A}$ to have identical vectors at each column. We can realize this by replicating the last output of the LSTM in Eq. (\ref{eq:lstm}) to every column of $\mathbf{A}$:
\begin{equation}
    \mathbf{A} = [\boldsymbol{\alpha}_L, \boldsymbol{\alpha}_L, ..., \boldsymbol{\alpha}_L]\in\mathbb{R}^{L\times L}
\end{equation}
where $\boldsymbol{\alpha}_L\in\mathbb{R}^L$ can be obtained using Eq. (\ref{eq:lstm}).

Now, to calculate the transformed curves using Eq. (\ref{eq:new_trans}), we need to know how to obtain both the window left indices $\mathbf{K}^T\mathbf{t}$ and the selected curve values $\mathbf{K}^T\mathbf{y}$ using the generated $\mathbf{D}$. We design a specific mask to help with the transformation between the window indicators and left index indicators. For an arbitrary column vector $\mathbf{x}$, the mask $\mathbf{M}_{\mathbf{x}}$ is defined as
\begin{equation}
    \mathbf{M}_{\mathbf{x}} =
    \begin{bmatrix}
    \text{Shift}(\mathbf{x}^T, 1)\\
    \text{Shift}(\mathbf{x}^T, 2)\\
    ...\\
    \text{Shift}(\mathbf{x}^T, L)\\
    \end{bmatrix} 
    \in\mathbb{R}^{L\times L}
\end{equation}
where $\text{Shift}(\mathbf{x}^T, j)=[0, ...,0, x_1, x_2, ..., x_{L-j}]\in\mathbb{R}^{1\times L}$ shifts the entries of $\mathbf{x}^T$ to the right by $j$ steps with 0s filling the left positions. 
We first calculate the two masks for both $\mathbf{t}$ and $\mathbf{y}$ as
\begin{equation}
    \mathbf{M}_{\mathbf{t}} =
    \begin{bmatrix}
    0 & 1 & 2 & ... & L-2 & L-1\\
    0 & 0 & 1 & ... & L-3 & L-2\\
    &&&...&&\\
    0 & 0 & 0 & ... & 1 & 2\\
    0 & 0 & 0 & ... & 0 & 1\\
    0 & 0 & 0 & ... & 0 & 0\\
    \end{bmatrix} 
    \in\mathbb{R}^{L\times L}
\end{equation}
and
\begin{equation}
    \mathbf{M}_{\mathbf{y}} =
    \begin{bmatrix}
    0 & y_1 & y_2 & ... & y_{L-2} & y_{L-1}\\
    0 & 0 & y_1 & ... & y_{L-3} & y_{L-2}\\
    &&&...&&\\
    0 & 0 & 0 & ... & y_1 & y_2\\
    0 & 0 & 0 & ... & 0 & y_1\\
    0 & 0 & 0 & ... & 0 & 0\\
    \end{bmatrix} 
    \in\mathbb{R}^{L\times L}
\end{equation}
Then, the window left indices $\mathbf{K}^T\mathbf{t}$ and the selected curve values $\mathbf{K}^T\mathbf{y}$ can be obtained by
\begin{equation}\label{eq:D_t}
    \mathbf{K}^T\mathbf{t} = (\mathbf{D}\odot\mathbf{M}_{\mathbf{t}})^T\mathbf{1}
\end{equation}
and
\begin{equation}\label{eq:D_y}
    \mathbf{K}^T\mathbf{y} = (\mathbf{D}\odot\mathbf{M}_{\mathbf{y}})^T\mathbf{1}
\end{equation}
where $\mathbf{1}=[1, 1, ..., 1]\in\mathbb{R}^L$ denotes the all-1 vector.
Finally, the transformed curve can be obtained by plugging Eq. (\ref{eq:D_t}) and (\ref{eq:D_y}) into Eq. (\ref{eq:new_trans}).

\section{Design Choice of Curve Encoder}

We selected the learning curve encoder from the commonly used sequential models such as LSTM, 1-dimensional CNN, multi-scale CNN~\cite{cui2016multi} (used in LCRankNet~\cite{wistuba2020learning}), and Temporal Convolutional Networks (TCN)~\cite{bai2018empirical}, and found that TCN works the best in modeling learning curves both effectively and efficiently.

\begin{figure}[t]
    \centering
    \includegraphics[width=0.55\linewidth]{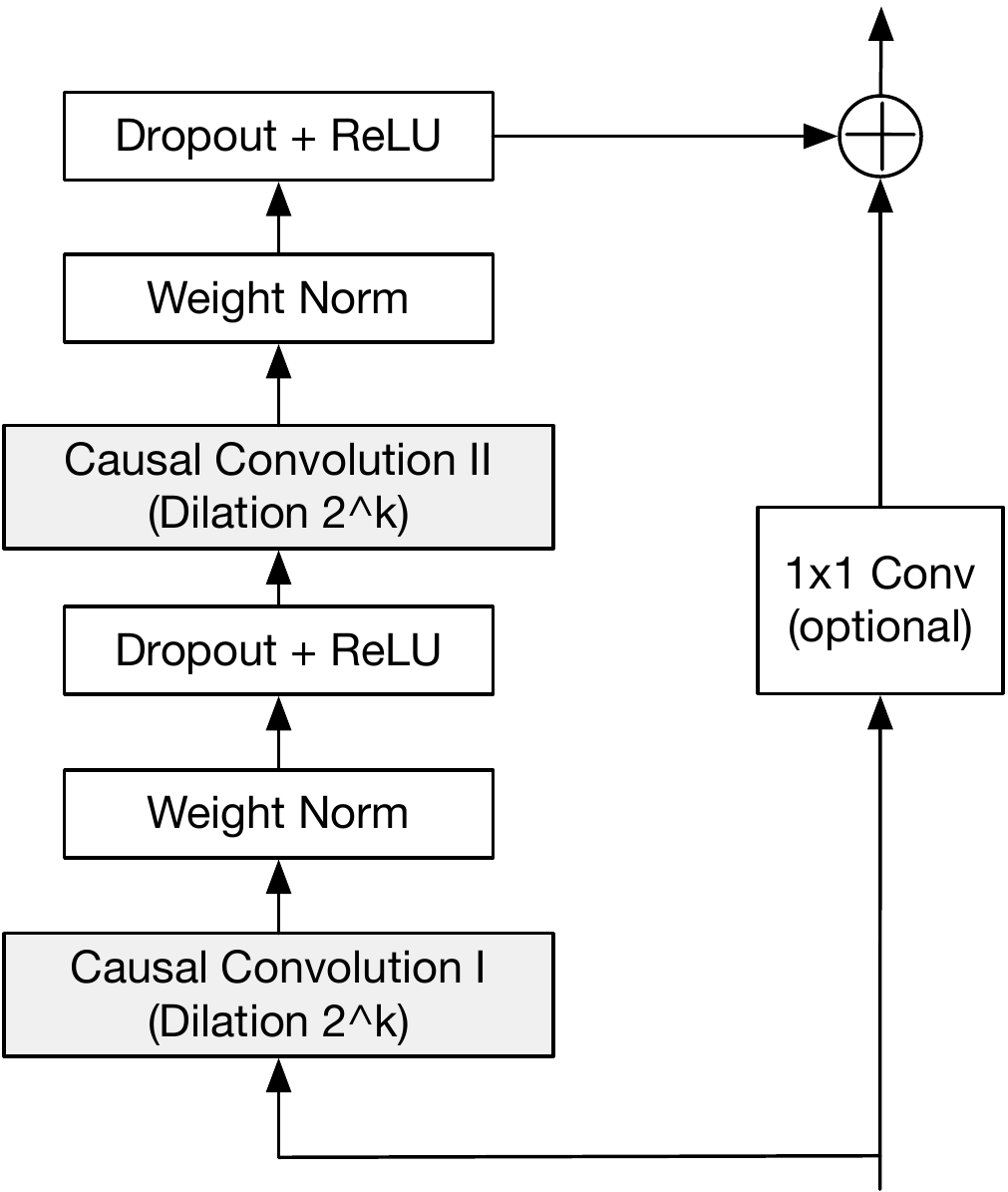}
    \caption{Architecture of TCN.}
    \label{fig:tcn}
\end{figure}

TCN is a type of efficient 1-dimensional convolutional neural network that processes sequential data by applying convolutions across time. TCN differs from typical 1-dimensional CNNs in its use of dilated causal convolution. Given an input sequence $\mathbf{X}=[\textbf{x}_1, ..., \textbf{x}_T]\in\mathbb{R}^{d\times T}$ and a convolution filter $\mathbf{f}\in\mathbb{R}^{k\times d}$, the dilated causal convolution operation $F$ on element $t$ of the sequence is defined as:
\begin{equation}
    F(\mathbf{x}_t) = (\mathbf{X} *_d \mathbf{f})(t) = \sum_{i=0}^{k-1} \textbf{f}_i^T\cdot \mathbf{x}_{t-d\cdot i}, \ \ s.t., \ t\geq k, \textbf{x}_{\leq 0}:= 0
\end{equation}
Where $d$ is the dilation factor, $k$ is the filter size, and $t - d\cdot i$ accounts for past values. Dilated convolution, using a larger dilation factor $d$, allows the output to represent a wider range of inputs, effectively increasing the receptive field of the convolution. Causal convolution, using only previous time steps in the convolution, ensures that no future information is used to process past data. This allows TCN models to have a similar directional structure to RNN models.

The output sequence $\textbf{X}'\in\mathbb{R}^{k\times T}$ of the dilated convolution layer can be written as:
\begin{equation}
    \textbf{X}' = [F(\mathbf{x}_1), F(\mathbf{x}_2), ..., F(\mathbf{x}_T)]
\end{equation}
TCN models are often regularized with Layer Normalization or Batch Normalization after the convolutional layer to improve performance. TCN models are typically composed of multiple causal convolutional layers with a wide receptive field that can handle long input sequences.

\end{appendices}

\end{document}